\documentclass[pdflatex,sn-mathphys-num]{sn-jnl}% Math and Physical 
\usepackage{float}
\usepackage{graphicx}%
\usepackage{multirow}%
\usepackage{amsmath,amssymb,amsfonts}%
\usepackage{amsthm}%
\usepackage{mathrsfs}%
\usepackage[title]{appendix}%
\usepackage{xcolor}%
\usepackage{textcomp}%
\usepackage{manyfoot}%
\usepackage{booktabs}%
\usepackage{algorithm}%
\usepackage{algorithmicx}%
\usepackage{algpseudocode}%
\usepackage{listings}%
\UseRawInputEncoding

\theoremstyle{thmstyleone}%
\theoremstyle{thmstyletwo}%

\theoremstyle{thmstylethree}%

\begin{document}

\title{TopView: Vectorising road users in a bird's eye view from uncalibrated street-level imagery with deep learning}


% \author*{\fnm{Manuscript submission}}

\author*[1]{\fnm{Mohamed R} \sur{Ibrahim}}\email{geomi@leeds.ac.uk}

\affil*[1]{\orgdiv{Institute of Spatial Data Science}, \orgname{University of Leeds}, \city{Leeds}, \country{UK}}


\abstract{Generating a bird's eye view of road users is beneficial for a variety of applications, including navigation, detecting agent conflicts, and measuring space occupancy, as well as the ability to utilise the metric system to measure distances between different objects. In this research, we introduce a simple approach for estimating a bird’s eye view from images without prior knowledge of a given camera’s intrinsic and extrinsic parameters. The model is based on the orthogonal projection of objects from various fields of view to a bird’s eye view by learning the vanishing point of a given scene. Additionally, we utilised the learned vanishing point alongside the trajectory line to transform the 2D bounding boxes of road users into 3D bounding information. The introduced framework has been applied to several applications to generate a live Map from camera feeds and to analyse social distancing violations at the city scale. The introduced framework shows a high validation in geolocating road users in various uncalibrated cameras. It also paves the way for new adaptations in urban modelling techniques and simulating the built environment accurately, which could benefit Agent-Based Modelling by relying on deep learning and computer vision.}




\maketitle

\section{Introduction}

Scene awareness across different views of a given scene represents an important subject not only in machine learning but also in studies related to understanding flows in cities and transports. Estimating a vectorised Bird-Eye View (BEV) representation of a given visual scene is useful for many real-world applications \cite{source1,source2}, including navigation, motion prediction \cite{source3}, robotics, or simply measuring distances and evaluating conflicts among road users whether it is a for understanding occupancy rates, social-distancing, accidents, or near misses. 

Understanding cities using computer vision, or more generally through machine learning, has gained the interest of planners and urbanists in the last few years \cite{source4}. However, the obstacles to combining both machine-based approaches and their outputs with the most frequent planning tool (maps) persist. If only the content of street-level imagery could automatically blend and localise to maps, this might benefit the utility of machine learning in cities at scale. Most recently, substantial progress has been made to create methods that can learn to generate a BEV representation for autonomous driving \cite{source1, source2}. However, most of these methods either tend to rely on a given camera’s parameters (intrinsic and extrinsic) for calibration making it limited for generalisation when cameras’ parameters are unknown \cite{source5,source6,source7,source8,source9,source10,source11}, or generate an image-based representation (i.e. image-to-image transition) \cite{source1,source12,source13} makes the yielding outcome useful only for few applications, excluding the ability to generate trajectories, or measuring distances without the need for an extra step of vectorising the raster output. Here we introduce a simple approach but a powerful one for generating a vector BEV representation from uncalibrated images which makes it applicable for both known and unknown cameras’ parameters, including internet data, and CCTV feeds whereas other current methods face shortcomings. The introduced approach, known as TopView, relies on learning the vanishing point of a given scene, while geometrically estimating the BEV vector space of a given space and the 3D representation of a given object from its 2D boxes estimated from the backbone of the model.

\begin{figure}[H]
  \centering

   \includegraphics[width=0.8\linewidth]{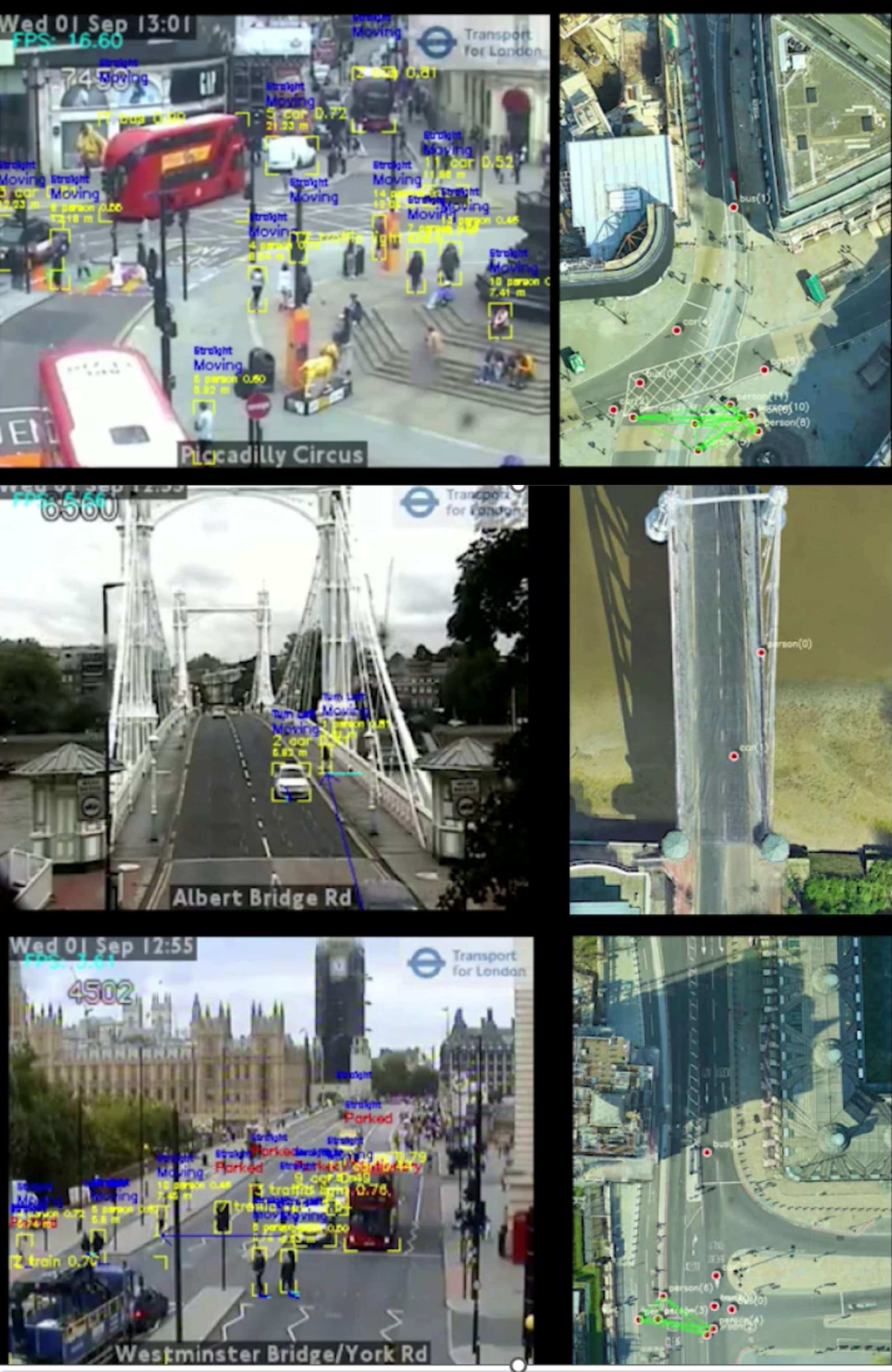}

   \caption{Examples of different sites of various street layouts and their estimated BEV map on a Google Map. }
   \label{fig:fig1}
\end{figure}

This study significantly extends the current methodologies used in computer vision for urban analytics by introducing a novel framework that supports robust and scalable analysis without the need for camera calibration. Our major contributions are detailed as follows: We propose a novel method for estimating Bird's Eye View (BEV) that operates independently of the camera's intrinsic and extrinsic parameters. This facilitates the application of BEV generation techniques to a broader range of uncalibrated images sourced from varied devices and viewpoints, thus making advanced visual analytics accessible in environments where camera calibration is impractical or unknown. Our approach simplifies the process of mapping and geo-locating road users' trajectories from uncalibrated 2D images to the geographic coordinate system through the innovative application of vanishing points to infer depth and scale. This significantly enhances the accuracy of object placement in virtual space, providing crucial data for traffic management and urban planning. Extending our method to video streams, we introduce a spatiotemporal representation of moving objects, encapsulating them as streams of tokens that capture dynamic changes over time. This provides a detailed and continuous narrative of object movements, which is invaluable for traffic flow analysis and surveillance. Additionally, we ensure the privacy of individuals by anonymising the representation of road users in both still images and video streams. This adherence to privacy laws and ethical standards makes our method suitable for sensitive environments where user consent may be unattainable. Finally, the practicality of our method is demonstrated through applications on diverse datasets, including CCTV footage from urban traffic systems (see Fig. \ref{fig:fig1}). These applications showcase the method's robustness across different settings and its capability to provide actionable insights for real-world challenges.

\section{Background}

We are not aware of any method for estimating BEV based solely on learning a vanishing point without knowing the cameras’ matrix or providing key points for a given perspective. However, our introduced method links with several knowledge domains: 

\subsection{Object detection }

Object detection is a cornerstone of computer vision with applications ranging from autonomous driving to security surveillance \cite{object_detect_survey1,object_detect_survey2, object_detect_survey3}. Traditionally, object detection relied on manual feature extraction combined with machine learning algorithms, using techniques like Histogram of Oriented Gradients (HOG) \cite{hog} and Scale-Invariant Feature Transform (SIFT) \cite{scale_invariant} alongside classifiers such as Support Vector Machines (SVM) \cite{svm}. However, the advent of deep learning has revolutionised the field, introducing more sophisticated and effective methods \cite{object_detect_survey1,object_detect_survey2, object_detect_survey3}.. Convolutional Neural Networks (CNNs) now dominate object detection, facilitating powerful feature extraction and recognition capabilities. Significant milestones include the development of Region-based CNNs (R-CNN) and its iterations \cite{fast_rcnn1,fast_rcnn2}, which efficiently localise and classify objects using region proposals. The You Only Look Once (YOLO) framework and its successors \cite{source35,source37,source36}) simplify detection into a single regression problem, enhancing the speed and feasibility of real-time applications. Similarly, the Single Shot MultiBox Detector (SSD) eliminates the need for proposal generation \cite{ssd}, directly predicting multiple bounding boxes and class probabilities, thus balancing speed with accuracy. These advancements in object detection pave the way for robust object localisation in bird's-eye view applications.

\subsection{Multi-view awareness based on homography }
In photogrammetry, moving from a given camera’s coordinates system to the world coordinate system is achieved by knowing the camera matrix including both intrinsic and extrinsic parameters \cite{source6, source9, source10, source11, source14, source15, source16}, as follows:
\begin{equation}
z_c \begin{pmatrix} u \\ v \\ 1 \end{pmatrix} = M \begin{pmatrix} x_w \\ y_w \\ z_w \\ 1 \end{pmatrix}, \text{ where } M = K[R \quad T]
\end{equation}
given that \( M \) represents the camera matrix, \( K \) and \( R \, T \) denote the intrinsic and extrinsic parameters of the camera respectively. They are defined as:
\begin{equation}
K = \begin{bmatrix} \alpha_x & \gamma & c_x  & 0\\ 0 & \alpha_y & c_y & 0\\ 0 & 0 & 1 & 0\end{bmatrix},  \text{ where } \alpha_x = f \cdot m_x \text{ and } \alpha_y = f \cdot m_y
\end{equation}
given that \( f \) represents the focal length of the camera in pixels, \( m_x \) and \( m_y \) represent the scale factors of relating pixels to distance, \( \gamma \) is the skew coefficient between the two axes of \( x \) and \( y \), which is often equal to zero. Lastly, \( c_x \) and \( c_y \) denote the principle point.

\begin{equation}
[R \quad T] = \begin{bmatrix} R_{3 \times 3} & T_{3 \times 1} \\ 0_{1 \times 3} & 1 \end{bmatrix}_{4 \times 4}
\end{equation}
Generally, extrinsic parameters represent the camera’s position and heading in the world coordinates, where \( T \) represents the origin position of the world coordinate system defined in the camera coordinate system, and \( R \) represents the rotation matrix of the camera. After calibrating cameras to world coordinates and by relying on homography, we can move from one camera pose to another as follows:
\begin{equation}
\begin{pmatrix} z_i x_i' \\ z_i y_i' \\ z_i \end{pmatrix} = H \begin{pmatrix} x_i \\ y_i \\ 1 \end{pmatrix},
\end{equation}
where \( \text{dst}(i) = (x_i', y_i') \), \( \text{src}(i) = (x_i, y_i) \), \( i = 0, 1, 2, 3 \)

given that \( \text{src} \) and \( \text{dst} \) represent the coordinates of the quadrangle vertices in the camera view and world coordinates respectively, \( (x_i, y_i) \) and \( (x_i', y_i') \) represent paired coordinate points in the camera and top-view planes respectively. Lastly, \( H \) represents the homography or the transformation matrix that is defined as:
\begin{equation}
H = \begin{bmatrix} h_{00} & h_{01} & h_{02} \\ h_{10} & h_{11} & h_{12} \\ h_{20} & h_{21} & h_{22} \end{bmatrix}
\end{equation}
where \( H \) is solved and calibrated by inputting the four paired points in the camera and top-view planes. Accordingly, by solving \( H \), the detected object in the camera plane can be transformed into the top-view plane.

Several studies have provided approaches with slight changes for image calibration based on this method \cite{source6, source7, source8, source9, source14, source16, source17, source18}. However, this method faces several shortcomings for automation and scalability such as 1) its requirement for calibrating cameras, limiting its usability to internet data, 2) the requirement for at least inputs of 4 points to represent the perspective to unwrap them in BEV image or corresponding points to estimate the transformation matrix, 3) Even when providing these points, without knowing the vanishing point, the BEV faces a high level of distortion for objects outside the bounds of the provided points.

\subsection{Geometric-based models for scene awareness }
Combining both features of Geometric constraints and machine learning, several methods have been achieved to generate a BEV map from a camera view \cite{source1, source2, source5, source19, source20} relied on a CNN model to obtain a homography matrix by transforming a monocular camera input to a BVP map. However, this approach lacks vectorising road users. However, these methods still require the camera’s model or several camera inputs lacking the ability of these models to apply directly to the ubiquitous uncalibrated images. 

\subsection{Multi-sensors fusion-based models }
Several methods have focused on estimating a BEV map based on fusing both RGB images and actual LiDAR data \cite{source1, source21, source22, source23} or pseudo-LiDAR generated from depth estimation \cite{source24}. This approach relies on generating a BEV map by encoding both data sources with early fusion or post-feature extraction to guide the model to learning orthogonal features.  For instance, \cite{source25} introduced a method for producing spatiotemporal Birds-Eye-View (BEV) representations from multi-camera footage and reasoning about multiple tasks collaboratively for vision-centric autonomous driving. \cite{source26} developed a model that learns the 3D representation of road users by fusing multiple camera inputs and extracting 2D and 3D feature streams based on the underlying geometric constraints in the BEV. While this approach has shown strength in localizing objects, it is limited to when LiDAR data is available, limiting its utility to certain applications primarily autonomous driving. Furthermore, it requires multiple calibration processes, as it requires not only calibrating the RGB images but also integrating the LiDAR data. 

\subsection{Image-to-Image translation }
While the objective of this research is to provide a vector representation rather than a raster one, it is worth mentioning that there have been several approaches that utilized image-to-image translation whether through adversarial learning or other approaches to generate an estimate of BEV map \cite{source1, source12, source13, source27}. These approaches included generating not only road users but also a semantic representation of the street layout \cite{source13, source28, source29} used encoder-decoder architecture to generate semantic segmentation for vehicle layouts from multiple camera sources. \cite{source30} developed a transformer-based model to extract the local road network layout in a BEV map based on a directed graph representation. \cite{source31} introduce a framework that includes a Hybrid Feature Transformation module that decouples learning and camera-based model approaches to output a semantic BEV map. \cite{source32} introduced a two-stage geometry-guided framework to generate a semantic BEV map from a monocular camera input. However, in practice, this approach, without geometric constraints, tends to provide noisy and unreliable outcomes when experimented with in unseen scenes, which we will report when comparing our results to existing methods.

\section{Methodology}

\begin{figure*}[H]
  \centering

   \includegraphics[width=1\linewidth]{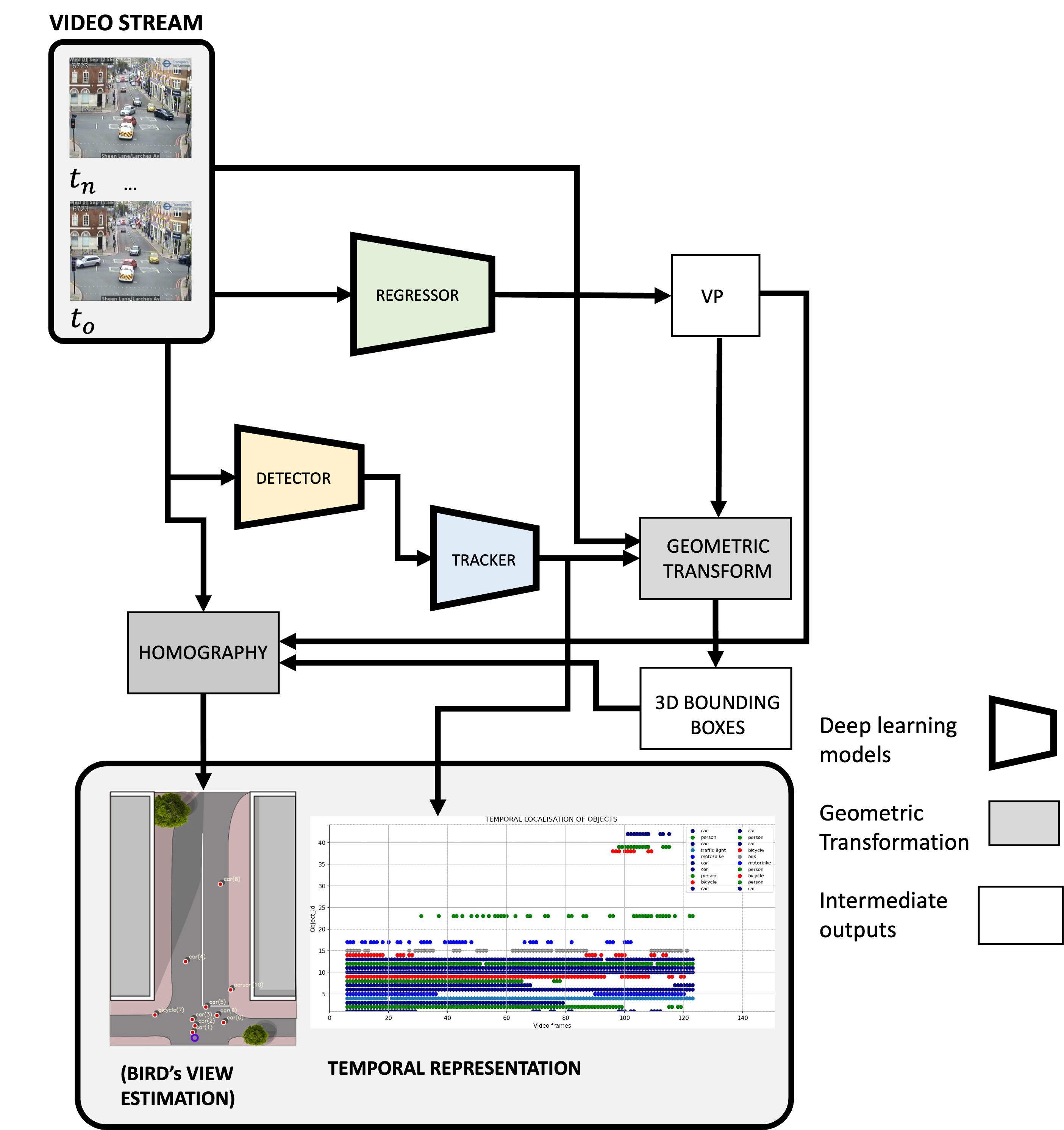}

   \caption{The overall architecture of temporal localisation and spatial representation of scene objects from video streams. }
   \label{fig:fig2}
\end{figure*}

\subsection{TopView framework}
Humans tend to navigate by knowing the relationship between objects and avoiding obstacles instead of knowing the exact depth of each point in a given scene. Here we present a framework, called TopView, to generate temporal and BEV representations of road users when feed with sequential images or BEV representation alone when feed with single images. The Topview framework only requires an image input without the need camera’s model which makes it scalable to different data sources when camera parameters are alone.  Fig. \ref{fig:fig2} shows the architecture of the overall framework. After a given input, the framework comprises five sub-models that output a vector BEV map of road users only in case of a given image, or a vector BEV map and temporal localisation of the tracked road users. First, the framework takes a given input to pass it through a deep model to regress the Vanishing Point (VP) and the horizon of a given scene. Afterwards, an object detector with a tracker system is utilised to localise road users. The tracked road users alongside the VP and the horizon line are passed through a Geometric transformation module that aims to transform the 2D bounding representation of road users into 3D bounding boxes. Last, all outcomes are based on a Homographic module to transform the data into the spatial representation of the BEV map alongside temporarily localising road users and finetuning the generated temporal paths to account for spatial occlusion and objects re-identification related issues such as miss-matched objects’ ids filtered based on the spatiotemporal patterns over multiple frame sequence. Besides the BEV map, the final output is a stream of paths representing road users’ trajectories of multi-dimensional information such as object id, type, 3D bounding boxes, and stationary status.

\textbf{VP model:} By relying on geometric principles, vanishing points are a well-known concept in 3D vision research for their ability to estimate 3D structures from 2D images \cite{source33,source34}. Accordingly, we developed a model to estimate vanishing points from natural uncalibrated scenes. To learn the vanishing point of a given scene, we trained a deep learning model that takes a given image to output the X and Y coordinates of its vanishing point. The model is built on the backbone of a truncated pre-trained model, and two additional branches of two Fully-Connected layers output each value of the coordinates of the vanishing point. The main reason for training our model is to ensure its accuracy and performance when used on the overall framework. 

\textbf{Object detection and tracking system:} We relied on YOLO architecture \cite{source35, source36}, in particular, we used a YoloV5m \cite{source37} as a backbone for TopView to detect road users including persons, cars, buses, trucks, bicycles, and motorbikes pre-trained on COCO dataset \cite{source38}, in which we find the results are optimal in terms of accuracy and speed in deployment. In the case of sequential frames are given, we used DeepSort architecture \cite{source39} to track objects, which based on Sort algorithms \cite{source40} coupled with a deep learning model to handle object occlusion. 

\begin{figure}[H]
  \centering

   \includegraphics[width=1\linewidth]{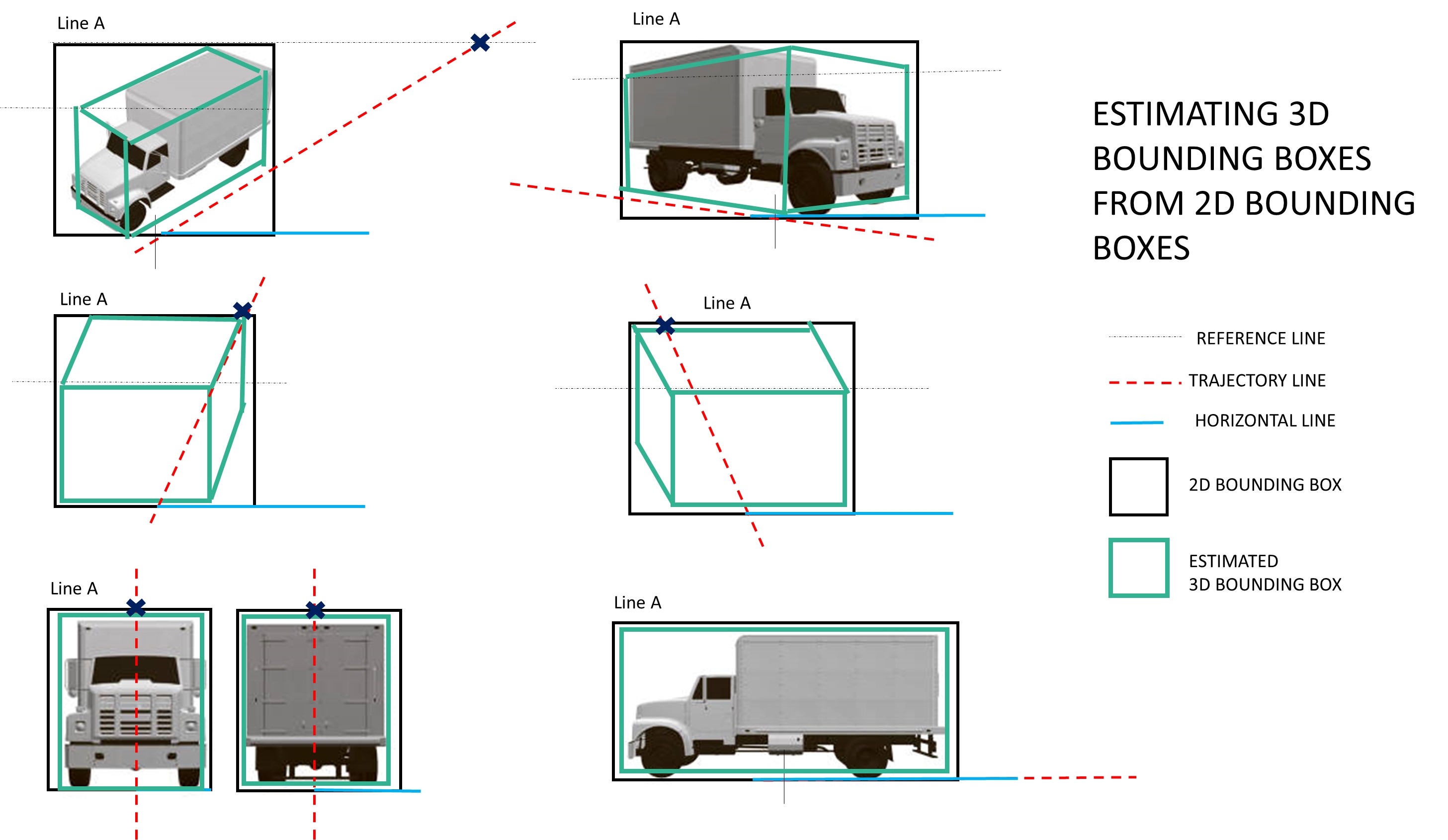}

   \caption{Estimating 3D bounding boxes from 2D bounding boxes from different poses of a given road user.}
   \label{fig:3d_bbox}
\end{figure}

\textbf{Geometric transformation:} To define the confined 3D bounding box within a 2D box, we used a simple geometric transformation that utilised both trajectory lines and the vanishing point to estimate the 3D bounding box for a given road user within a scene. Fig. \ref{fig:3d_bbox} shows a few examples of different poses of a given object and the potential representation of the 3D bounding box that belongs to a given motion pose. We are aware of the efforts in the literature that aimed towards learning to estimate the 3D bounding box from the 2D bounding box of a given object \cite{source24, source41, source42, source43, source44, source35, source46}. However, despite learning the 3D representation from 2D representation, this method is still a given camera’s model limiting its utility to other data sources. Here we show that we can achieve the same output with a simple Geometric transformation between both scenes, utilising scene information such as VP that we already automated and trajectory lines and therefore we will not need a given camera’s model as presented by the learned method.  

Algorithm 1 shows the algorithm for transforming 2D bounding boxes to 3D bounding boxes by heuristically estimating and bounding the 6DOF of a 3D bounding box in the given 2D bounding box. The algorithm estimates the geometry of a given 3D box by understanding the orientation of a given road user. This can be estimated by understanding the relationship between a given object's trajectory line, horizon, and the reference line derived from the location of the vanishing point in a given scene. The variables used in the algorithm are as follows: $Q_{\text{trajectory\_line}}$ is the set of points that form the trajectory line of the moving object. Each point in the set is represented as $q$. The vanishing point in the image, represented by its coordinates $(vp_x, vp_y)$, is denoted as $vp$. The width of the image is represented by $w$. The coordinates that define the corners of the 2D bounding box around the detected object are $x_1, x_2, y_1, y_2$, where $x_1$ and $x_2$ are the $x$-coordinates of the left and right edges of the 2D bounding box, respectively, and $y_1$ and $y_2$ are the $y$-coordinates of the top and bottom edges, respectively. The variable $F$ represents the estimated orientation of the car (moving object) and is initially set to ``Undefined.'' The top edge of the 2D bounding box, $L_A$, is represented by the two points: $(x_1, y_1)$ and $(x_2, y_1)$. The midpoint of the top edge, denoted as $M$, is calculated as $M = \frac{(x_1 + x_2)}{2}$. The algorithm checks for intersections between points on the trajectory line and the top edge $L_A$ to determine the object's orientation relative to the midpoint $M$ and the vanishing point $vp$. If no intersection is found, the orientation $F$ is set to ``side view.'' Based on these calculations and conditions, the estimated car orientation $F$ and the 3D bounding box are the outputs. Our approach is effective in creating reliable and explainable bounding boxes inside the scene for non-stationary objects even without learning. This approach can also be utilised for stationary objects by replacing the trajectory line with the edge line of a given object; however, we leave this for future investigation.

\begin{algorithm}
\caption{Estimation of Car Orientation and 3D Bounding Box from Trajectory Lines}
\label{alg:car_orientation_3d_bounding_box}
\begin{algorithmic}[1]
\State \textbf{Input:} $Q_{\text{trajectory\_line}}$ set of trajectory line points $q$
\State \textbf{Input:} $vp$ vanishing point in image coordinates $(vp_x, vp_y)$
\State \textbf{Input:} $w$ width of the image
\State \textbf{Input:} $x_1, x_2, y_1, y_2$ coordinates of the 2D bounding box
\State $F \gets \text{Undefined}$ \Comment{Initialize car orientation}
\State $L_A \gets \left[ (x_1, y_1), (x_2, y_1) \right]$ \Comment{Top edge of the bounding box}

\ForAll{$q \in Q_{\text{trajectory\_line}} \text{ such that } q \cap L_A \neq \emptyset$}
    \State $(q_x, q_y) \gets q \cap L_A$ \Comment{Intersection point of trajectory line with $L_A$}
    \State $M \gets \frac{x_1 + x_2}{2}$ \Comment{Midpoint of $L_A$}
    \If{$\left| vp_x - \frac{w}{2} \right| \approx 0$} \Comment{Check if vanishing point is at image center}
        \If{$q_x < M$}
            \State $F \gets \text{``turning left''}$
        \ElsIf{$q_x > M$}
            \State $F \gets \text{``turning right''}$
        \Else
            \State $F \gets \text{``moving straight''}$
        \EndIf
    \Else
        \If{$q_x < M - \left| vp_x - \frac{w}{2} \right|$}
            \State $F \gets \text{``turning left''}$
        \ElsIf{$q_x > M - \left| vp_x - \frac{w}{2} \right|$}
            \State $F \gets \text{``turning right''}$
        \Else
            \State $F \gets \text{``moving straight''}$
        \EndIf
    \EndIf
\EndFor

\If{$F = \text{Undefined}$}
    \State $F \gets \text{``side view''}$ \Comment{Case when no intersection is found}
\EndIf

\State \textbf{Output:} $F$ \Comment{Estimated car orientation}
\State \textbf{Output:} 3D bounding box \Comment{Based on $F$ and 2D box dimensions}
\end{algorithmic}
\end{algorithm}

% \begin{algorithm}
% \caption{Detecting Car Orientation and 3D Bounding Box Based on Trajectory Line}
% \begin{algorithmic}[1]
% \State \textbf{Input:} $q \in Q_{\text{trajectory\_line}}, vp, w$
% \State \textbf{Input:} $x_1, x_2, y_1, y_2 \Rightarrow O_{\text{boundingbox}}$
% \State $F \approx \text{car\_orientation}$
% \State $L_A \gets \left[ (x_1, y_1), (x_2, y_1) \right]$

% \ForAll{$q \in N \wedge q \cap L_A$}
%     \If{$\left| vp_x - \frac{w}{2} \right| \approx 0$}
%         \If{$(q \cap L_A)_x < \frac{x_1 + x_2}{2}$}
%             \State $F \gets \text{``turning left''}$
%         \ElsIf{$(q \cap L_A)_x > \frac{x_1 + x_2}{2}$}
%             \State $F \gets \text{``turning right''}$
%         \ElsIf{$(q \cap L_A)_x \approx \frac{x_1 + x_2}{2}$}
%             \State $F \gets \text{``moving straight''}$
%         \EndIf
%     \Else
%         \If{$(q \cap L_A)_x < \left| \frac{x_1 + x_2}{2} - \left| vp_x - \frac{w}{2} \right| \right|$}
%             \State $F \gets \text{``turning left''}$
%         \ElsIf{$(q \cap L_A)_x > \left| \frac{x_1 + x_2}{2} - \left| vp_x - \frac{w}{2} \right| \right|$}
%             \State $F \gets \text{``turning right''}$
%         \Else
%             \State $F \gets \text{``moving straight''}$
%         \EndIf
%     \EndIf
% \Else
%     \State $F \gets \text{``side view''}$
% \EndFor

% \State \textbf{Output:} $F$
% \State \textbf{Output:} 3D bounding box
% \end{algorithmic}
% \end{algorithm}

\begin{figure}[H]
  \centering

   \includegraphics[width=1\linewidth]{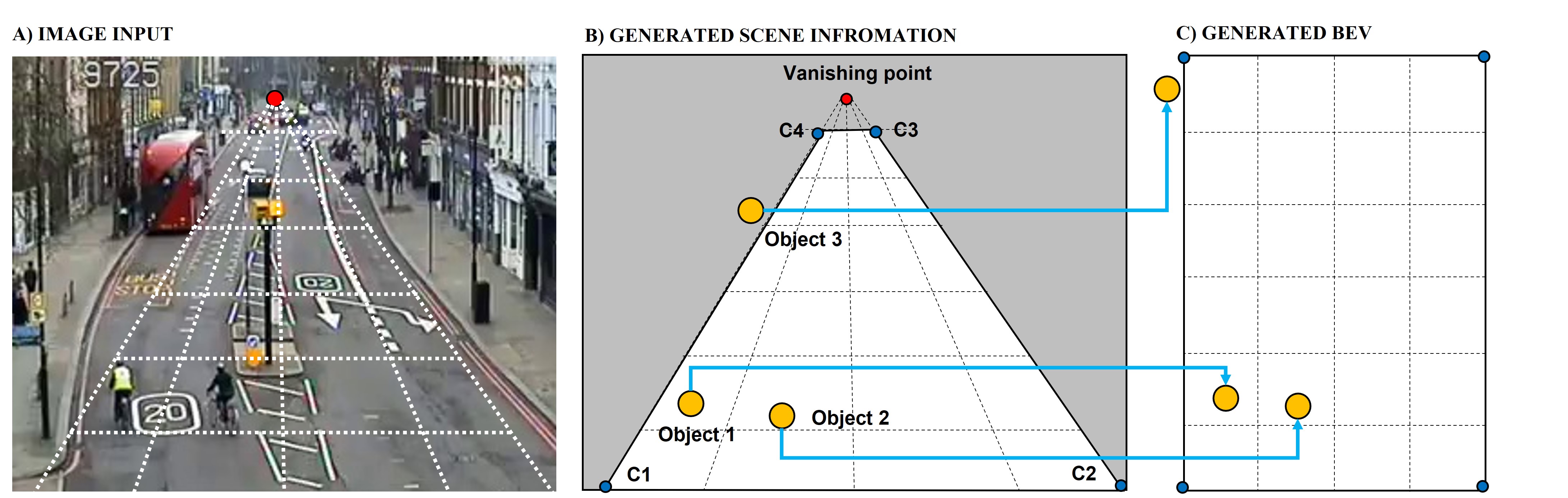}

   \caption{Transforming a given image input to a Bird’s eye view.}
   \label{fig:homography}
\end{figure}

\textbf{Homography:} To estimate a perspective plane grid, we create a horizontal line at the bottom of a given scene that is evenly subdivided by several points, and we utilise the detected VP to draw several lines from this VP to the bottom of each point at the above-mentioned horizontal line. We employed the four intersection points generated by these radial lines originating by the VP point and the upper and lower horizontal lines inside the confines of a specific image's VP and lower half. As a result, we developed an automated four-point representation of a particular scene in which they match a corresponding four-point rectangle representation of the BEV map's vector space. We applied homography to connect road users' point coordinates in a particular image plane to the newly estimated vector space of the BEV map (See Fig. \ref{fig:homography}).

\subsection{Objective Loss and Evaluations}
For the VP model, we trained the model based on the logcosh loss function for each coordinate of the point. For small values of \( x \) and the big one, respectively, \(\log(\cosh(x))\) is roughly equivalent to \((x^2) / 2\) and \(|\,x\,| - \log(2)\). Consequently, the logcosh function is mostly like the mean squared error while being less sensitive to the rare extremely inaccurate prediction.

For Object localisations, the objective loss is defined based on the weighted sum of the localization loss (\(L_{loc}\)) and confidence loss (\(L_{conf}\)) for the introduced backbone of object detection to detect and localise humans as follows:
\begin{equation}
L(x, l, g) = \frac{1}{N} (L_{conf}(x, c) + \alpha L_{loc}(x, l, g))
\end{equation}
By cross-validating the model, the loss is set to 0 if \(N=0\) and \(\alpha\) is set to 1 given that \(N\) is the default bounding box. Based on a Softmax loss for each class, the confidence loss is a cross-entropy loss (\(c\)). The default bounding box's centre (\(cx, xy\)), as well as its width (\(w\)) and height (\(h\)), establish the parameters of the predicted box (\(l\)), and the localization loss is defined as a smooth loss between those parameters and the ground truth bounding box (\(g\)). It is defined as follows:
\begin{equation}
L_{loc}(x, l, g) = \sum_{i \in Pos}^{N} \left[ \sum_{m \in \{cx, xy, w, h\}} x_{ij}^{k} \text{smooth} L_1(l_i^m - \hat{g}_i^m) \right]
\end{equation}
Evaluating a BEV map in a new dataset remains a challenge which poses an open question in the literature. Even though we have taken a heuristic approach to generate a BEV map, we evaluate the relationship between the different mapped objects after calibrating the image to its geolocation. We used Google Maps as a qualitative measure for verifying the localisations of the objects from the image plane to the real-world coordinate. 

\subsection{Implementations} 

\textbf{Data processing:} Each image used for training the VP model was normalised and downsized to a $(300 \times 300)$ grayscale image. To ensure the model focused on the geometric structures within the image, we applied the Canny edge detector filter to all images, preserving the edge details. The filtered images were then fed into the VP model for further processing.

\textbf{VP Model training:} We employed a MobileNet architecture \cite{source47}, pre-trained on ImageNet, as the backbone for our VP model. This involves removing the fully connected layers from the pre-trained MobileNet. We then applied a Global Average Pooling layer to determine the X and Y coordinates of the vanishing point. The network was extended with two separate branches, each designed to predict one of the coordinates (X or Y). Each branch consisted of two Fully-Connected layers with 100 neurons each, activated by a ReLU function to ensure non-linearity. To prevent overfitting, a dropout layer with a dropout rate of 0.5 was added after each Fully-Connected layer. The final output layer of each branch consisted of a single neuron activated by a linear function. This architecture was trained for 100 epochs with a batch size of 256, using the Adam optimizer \cite{source48}. An early stopping callback was utilized to halt the training process if the model's loss did not improve for 5 consecutive epochs, thereby optimizing training time and improving the model's generalization capability.

\textbf{Object detection and tracking system:} For object detection, we used the YOLOv5m \cite{source37} architecture, pre-trained on the COCO dataset \cite{source38} for detecting several classes of road users, including pedestrians, cars, buses, trucks, bicycles, and motorbikes. Given sequential frames as input, object tracking was achieved using the DeepSORT algorithm \cite{source39}, which combines the SORT algorithm for data association based on bounding box overlaps, and a deep appearance descriptor to maintain object identities during occlusion. This tracking mechanism is crucial for analysing the temporal localisation and movement patterns of objects.

\textbf{Geometric transformation:} The transformation from 2D to 3D bounding boxes leverages the geometric relationship between the object's position, its trajectory line, and the vanishing point. This transformation is automated using the algorithm described in Algorithm 1. In essence, the algorithm adjusts the 3D bounding box to fit within the 2D bounding box, derived from the predicted trajectory line and vanishing point, ensuring minimal error in the computed 3D orientation and dimensions.

\textbf{Homography:} We automated the computation of the homography matrix to map points from the image plane to the Bird’s Eye View (BEV) vector space. This was done by delineating a horizontal line at the bottom of the image, subdividing it evenly, and drawing lines from the vanishing point to each of these subdivisions. We identified four intersection points between these lines and upper and lower horizontal lines, using them for the homography transformation. These points represent the bounding quadrilateral in the image plane, which was mapped to a corresponding rectangle representing the BEV space. This transformation facilitated the precise localisation of road users in the BEV map.

\subsection{Materials and Experiments}
We trained the model for inferring the vanishing point in a given scene by combining multiple open-access data sources to ensure the diversity of outdoor scenes. These datasets include six different types of datasets: 1) London Streetview \cite{source49}, 2) Boston Streetview \cite{source49}, 3) Norway Streetview \cite{source49}, 4) Flickr \cite{source50}, 5) AVA \cite{source50} and 6) TfL London CCTV \cite{source51}. The combined dataset comprises 172,576 images. The variety of these datasets ensures that the model sees data at different times of the day (for instance, Google Streetview data is only day-time whereas the rest contains night-time dataset), different weather, and different fields of view. We randomly divided the dataset into training, validation, and testing groups in the following ratios: 80\%, 10\%, and 10\%.

To further report on the robustness of our orthogonal-based approach to the overall framework, we implemented and tested our model in sequential datasets by utilising London CCTV video streams to verify geolocational references of objects in a given scene in a Google map. Nevertheless, we also apply our approach to a sample of internet data to scale its validity.

\begin{table}[ht]
\centering
\caption{Datasets for Training VP Model}
\label{tab:datasets_vp_model}
\begin{tabular}{@{}lc@{}}
\toprule
\textbf{Dataset} & \textbf{Number of Images} \\
\midrule
London Streetview & 46,281 \\
Boston Streetview & 38,215 \\
Norway Streetview & 84,447 \\
Flickr & 959 \\
AVA & 1,315 \\
TfL London CCTV & 1,359 \\
\bottomrule
\end{tabular}
\end{table}

\section{Results}

\textbf{Performance Across Different Scenarios: } We assessed the robustness of the proposed framework under various conditions, including different lighting environments and weather scenarios. The results of this evaluation are summarised in Table~\ref{tab:robustness_evaluation}. We utilise mean Average Precision (mAP) to evaluate the accuracy of object detection and Mean Squared Error (MSE) to measure the accuracy of vanishing point (VP) estimation. Table \ref{tab:robustness_evaluation} illustrates that the framework performs exceptionally well under daylight conditions, achieving an object detection mAP of 90.1\% and a VP estimation MSE of 0.038. These results demonstrate the system's high accuracy and low error in optimal lighting conditions. However, in nighttime scenarios, the performance slightly decreases, achieving an object detection mAP of 85.7\% and a VP estimation MSE of 0.045. The decrease is expected due to the challenges presented by low-light environments. In adverse weather conditions, the framework faces the most significant challenges, with the object detection mAP dropping to 82.3\% and the VP estimation MSE increasing to 0.060. These variations indicate the impact of environmental factors on detection and estimation accuracy. Overall, while the framework exhibits robustness across different scenarios, there is a noticeable performance decline in less favourable conditions, emphasising the need for algorithms that can adapt to such variability.

\begin{table*}[ht]
\centering
\caption{Robustness Evaluation Across Different Scenarios}
\label{tab:robustness_evaluation}
\begin{tabular}{@{}lcc@{}}
\toprule
\textbf{Scenario} & \textbf{Object Detection mAP (\%)} & \textbf{VP Estimation MSE} \\
\midrule
Daylight & 90.1 & 0.038 \\
Nighttime & 85.7 & 0.045 \\
Adverse Weather & 82.3 & 0.060 \\
\bottomrule
\end{tabular}
\end{table*}

\begin{figure}[H]
  \centering

   \includegraphics[width=1\linewidth]{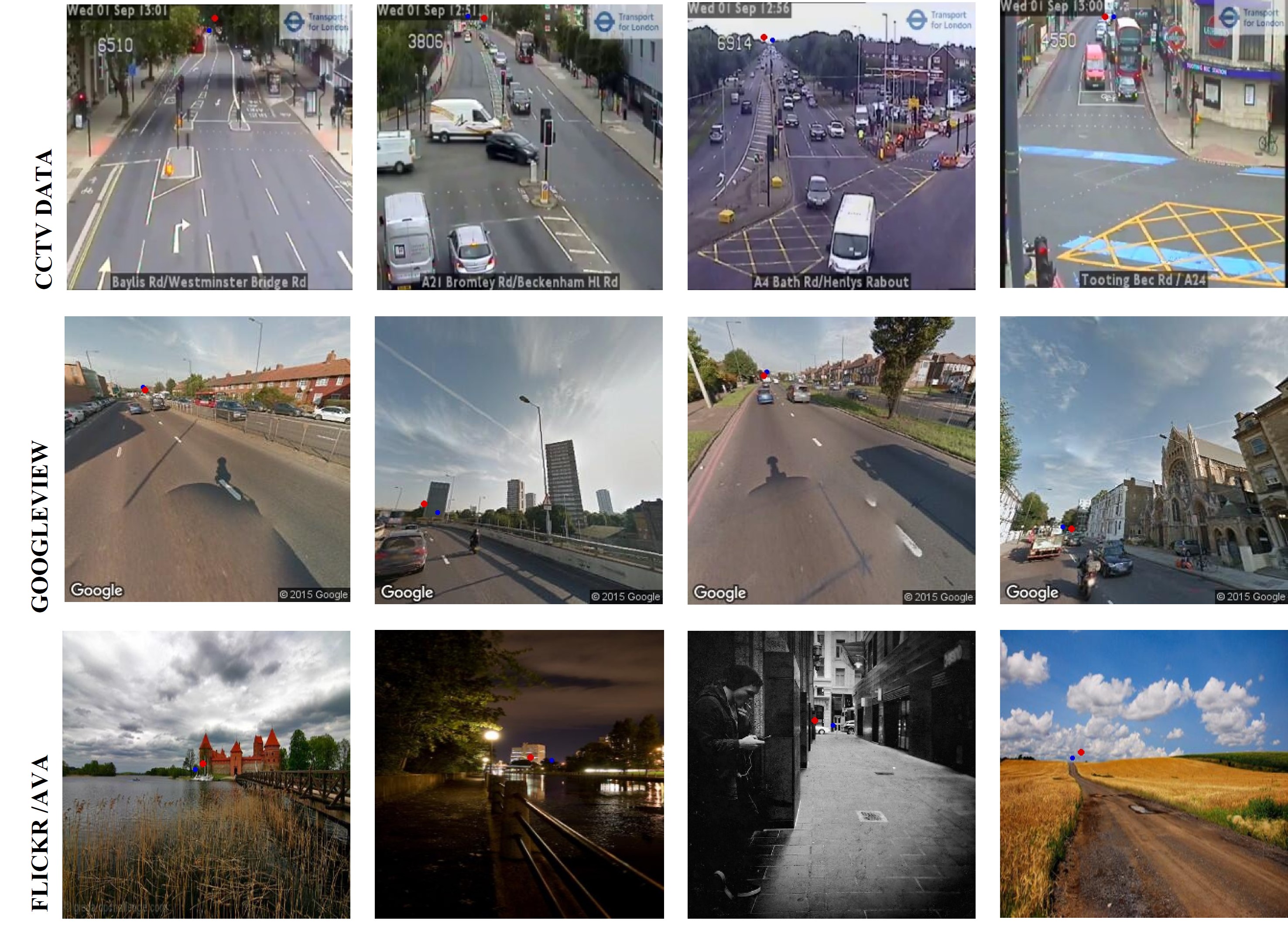}

   \caption{Sample of estimated VP points in the various dataset types (Predicted point in red dot, ground truth in blue dot). a given image input to a Bird’s eye view.}
   \label{fig:vp_examples}
\end{figure}

\textbf{Estimating vanishing points:} After training the VP model, Fig. \ref{fig:vp_examples} shows a sample of the predicted vanishing point (in red) and the ground truth one (in blue) in a variety of images with varying lighting conditions and fields of view obtained from various data sources. Despite the complexity of the presented scene layouts and their varying conditions, the trained model demonstrates good validation in grasping the orthogonal structure of a given scene and recognising its vanishing point. Based on these observed scenarios, there is still a small margin of error between the predicted and ground truth values of the vanishing points, particularly for the vanishing points' X coordinates.

\begin{figure}[H]
  \centering

   \includegraphics[width=1\linewidth]{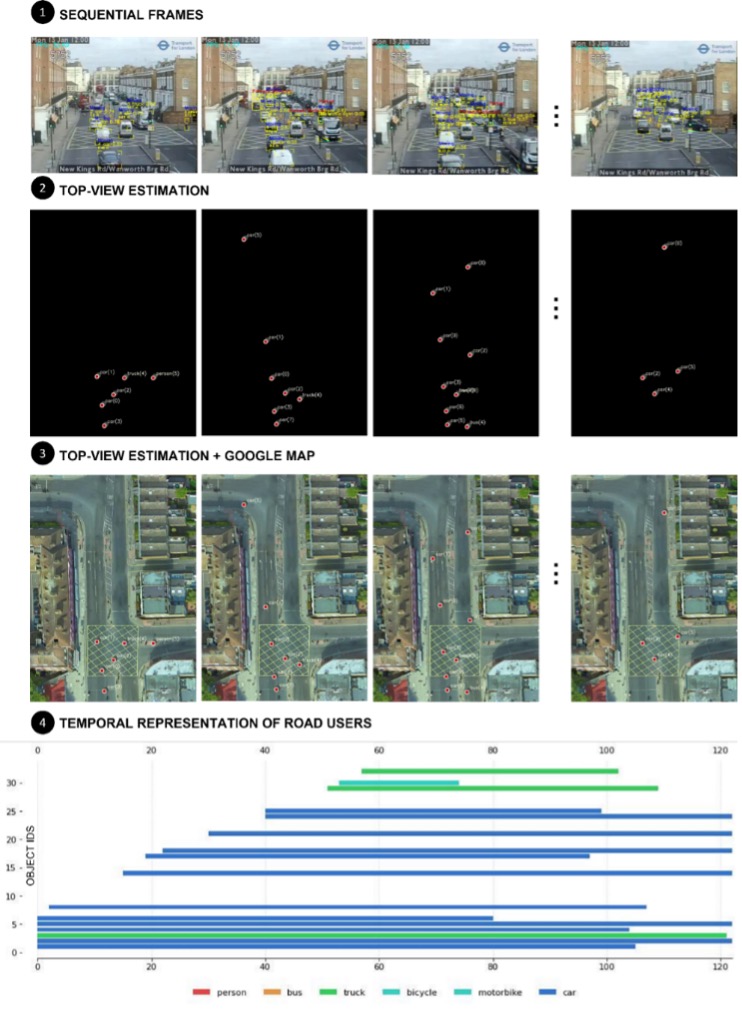}

   \caption{Transforming road users in London CCTV camera to a Bird’s eye estimation with Google Maps to verify the geolocation of road users.}
   \label{fig:fig6}
\end{figure}

\begin{figure}[H]
  \centering

   \includegraphics[width=1\linewidth]{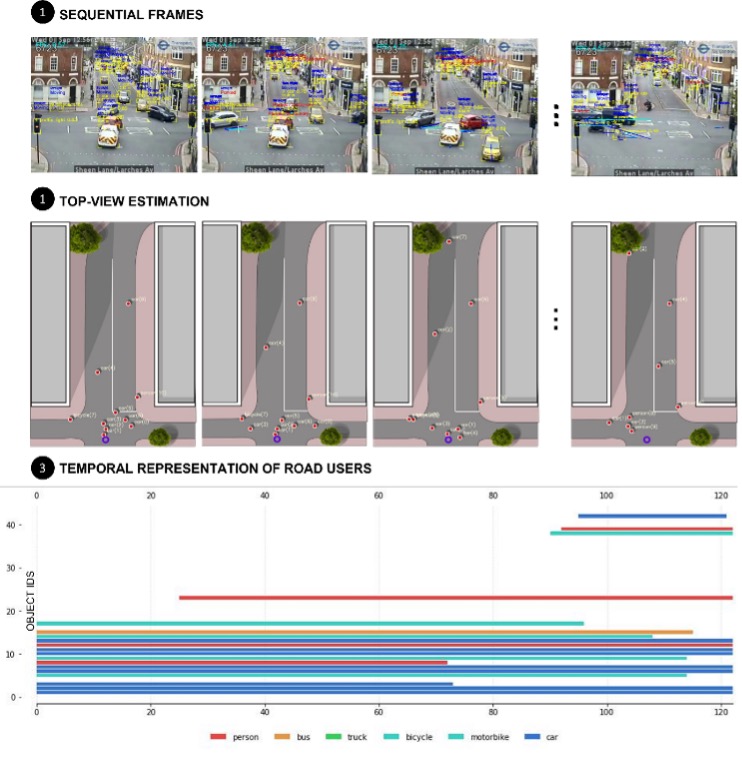}

   \caption{Transforming road users in London CCTV camera to a Bird’s eye estimation with a Semantic map to gamify a given video input. }
   \label{fig:fig7}
\end{figure}

\textbf{Localising road users in a BEV map: }
Fig. \ref{fig:fig6} shows an example of estimating a BEV map for a given video file from TfL London CCTV data. It shows first the transition of video frames to BEV maps with and without Google Maps. It shows a highly accurate localisation when qualifying the patterns of road users in Google Maps. It also shows the temporal localisations of road users as tokens, highlighting the appearance and disappearance through the time interval of the given video file. Furthermore, Fig.  \ref{fig:fig7} shows another scene from the dataset, with a semantic representation of the street layout. This scene exemplifies the accuracy of localising a variety of road users, such as the bicycle on the pavement (left-hand side) and the pedestrian on the pavement (right-hand side), as well as the complexity of the many vehicles at the road junction, taking into account their stationary state. It is worth mentioning that we only highlight the semantic segmentation here without showing or assessing how to generate a given street layout, leaving it for future investigation on how to use this approach for the gamification of London CCTV video streams. In doing so, this gamified approach could be useful for several studies and modelling techniques, particularly agent-based modelling and data assimilation while protecting individual road users' privacy.

\begin{figure}[H]
  \centering

   \includegraphics[width=1\linewidth]{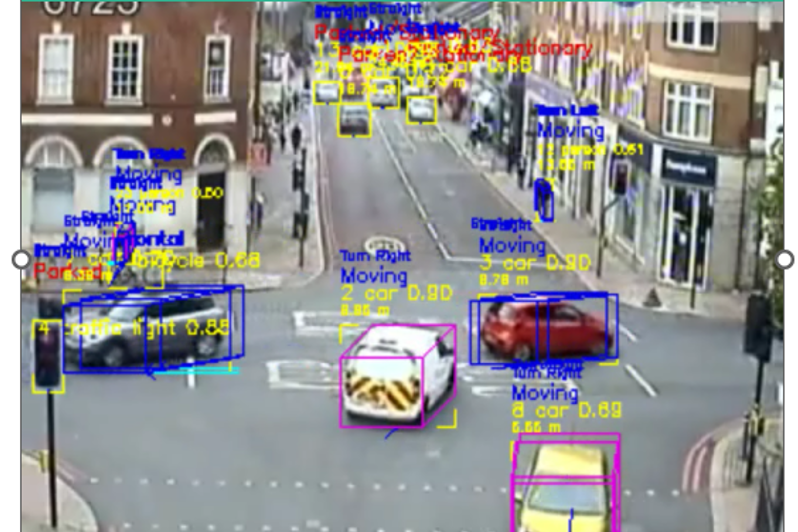}

   \caption{An example of the estimated 3D bounding box based on the introduced Algorithm.}
   \label{fig:3d_example}
\end{figure}

\textbf{3D bounding box estimation: } Fig. \ref{fig:3d_example}  shows an example of how a 3D bounding box can be effectively inferred, deterministically without prior learning, based on geometry and the orientation of road users in a given scene, using only the introduced algorithm that constrains 3D bounding boxes based on a given input of a 2D bounding box, a trajectory line, and a vanishing point. The diagram also depicts road users' orientation and stationary state.

% comparing results
%%%%%%%%%%%%%%%%%%%%%%%%%%%%%%%%%%%%%
%%%%%%%%%%%%%%%%%%%%%%%%%%%%%%%%%%%%%

\subsection{Comparing the results with state-of-the-art methods}
The comparison evaluates various Bird’s Eye View (BEV) generation methods across multiple dimensions, including their ability to localise road users, perform object tracking, and generate 3D bounding boxes.
Table \ref{qualitative_comparison_table} and Table \ref{comparison_sota_3D} present a qualitative and quantitative comparison of various BEV generation methods. The qualitative table (Table \ref{qualitative_comparison_table}) highlights key features of each method, such as the need for camera calibration, object classes detectable, ability to generate 3D bounding boxes, tracking capabilities, and whether the method produces vector data. The "TopView" framework, proposed in this paper, stands out by not requiring camera calibration and supporting a wide range of object classes, including pedestrians, cars, and bicycles. It can generate 3D bounding boxes through geometric transformation and includes temporal tracking capabilities, producing vector data suitable for multiple applications. In contrast, methods like Geometry-based Homography and Learned Depth Estimation require camera calibration and are limited in their object class detection and tracking capabilities. The Multi-Sensor Fusion approach, although capable of producing 3D bounding boxes, relies on additional data sources such as LiDAR.The quantitative comparison table (Table \ref{comparison_sota_3D}) presents metrics such as average translation error (mATE), average scale error (mASE), average orientation error (mAOE), average velocity error (mAVE), and average attribute error (mAAE), as well as NuScenes Detection Score (NDS) and mean average precision (mAP) for the different methods. The "TopView" framework shows competitive performance across various metrics. Other methods, such as CenterFusion and VoxelNet, are also compared to highlight their effectiveness in 3D object detection on the nuScenes validation set. These tables collectively provide a comprehensive comparison of the methods, highlighting the strengths and limitations of each and positioning the "TopView" framework as a versatile and effective approach for BEV generation in uncalibrated street-level imagery.

\begin{table}[ht]
\centering
\caption{Qualitative Comparison of BEV Generation Methods. Modality: 1 = Camera, 2 = LiDAR, 1 \& 2 = Both. All objects includes Pedestrian, Cars, Bicycles}
\label{qualitative_comparison_table}
\begin{tabular}{@{}lcccccc@{}}
\toprule
\textbf{Method}  & \textbf{Modality} & \textbf{Calib. Req.} & \textbf{Obj. Classes} & \textbf{3D BBox} & \textbf{Tracking} & \textbf{Vector Data} \\ 
\midrule
OFT-Net (BEV) \cite{source5}             & 1 & Yes & Vehicles only & No & No & No \\
Image-to-Image \cite{source12}           & 1 & No & None & No & No & No \\
PillarFlow  \cite{source22}              & 2 & Yes & Vehicles & Yes & Yes & No \\
BEVerse \cite{source25}                  & 1 & Yes & Vehicles & Yes & No & No \\
GitNet \cite{source32}                   & 1 & No & All & No & No & No \\
CenterFusion \cite{nabati2021centerfusion} & 1 \& 2 & Yes & All & Yes & Yes & Yes \\
VoxelNet \cite{maturana2015voxnet}       & 2 & Yes & All & Yes & Yes & Yes \\
PointPillar \cite{lang2019pointpillars}  & 2 & Yes & All & Yes & Yes & Yes \\
CenterNet \cite{zhou2019objects}         & 1 & Yes & All & Yes & Yes & Yes \\
FCOS3D \cite{wang2021fcos3d}             & 1 & Yes & All & Yes & Yes & Yes \\
DETR3D \cite{wang2021detr3d}             & 1 & Yes & All & Yes & Yes & Yes \\
PGD \cite{wang2022pgd}                   & 1 & Yes & All & Yes & Yes & Yes \\
PETR-R50 \cite{liu2022petr}              & 1 & Yes & All & Yes & Yes & Yes \\
PETR-R101 \cite{liu2022petr}             & 1 & Yes & All & Yes & Yes & Yes \\
PETR-Tiny \cite{liu2022petr}             & 1 & Yes & All & Yes & Yes & Yes \\
BEVDet-Tiny \cite{bevdet2022}            & 1 & Yes & All & Yes & Yes & Yes \\
BEVDet-Base \cite{bevdet2022}            & 1 & Yes & All & Yes & Yes & Yes \\

\textbf{TopView (Proposed)}              & 1 & No & All & Yes & Yes & Yes \\

\bottomrule
\end{tabular}
\end{table}

\begin{table}[ht]
\centering
\caption{Comparison with the state-of-the-art methods for 3D object detection on the nuScenes \cite{caesar2020nuscenes} validation set. Modality: 1 = Camera, 2 = LiDAR, 1 \& 2 = Both.}
\label{comparison_sota_3D}
\begin{tabular}{@{}lccccccccc@{}}
\toprule
\textbf{Method}      & \textbf{Modality}  & \textbf{mATE ↓} & \textbf{mASE ↓} & \textbf{mAOE ↓} & \textbf{mAVE ↓} & \textbf{mAAE ↓} & \textbf{NDS ↑} & \textbf{mAP ↑} \\ 
\midrule
CenterFusion \cite{nabati2021centerfusion}    & 1 \& 2             & 0.540           & 0.142           & 0.649           & 0.263           & 0.535           & 0.453          & 0.332          \\
VoxelNet \cite{maturana2015voxnet}       & 2                  & 0.292           & 0.253           & 0.316           & 0.328           & 0.306           & 0.716          & 0.264          \\
PointPillar \cite{lang2019pointpillars}     & 2                  & 0.295           & 0.803           & 0.268           & 0.511           & 0.374           & 0.303          & 0.860          \\
CenterNet \cite{zhou2019objects}       & 1                  & 0.321           & 0.818           & 0.188           & 0.326           & 0.191           & 0.330          & 0.183          \\
FCOS3D \cite{wang2021fcos3d}         & 1                  & 0.285           & 0.935           & 0.200           & 1.242           & 0.361           & 0.311          & 0.751          \\
DETR3D \cite{wang2021detr3d}         & 1                  & 0.303           & 0.794           & 0.216           & 1.152           & 0.356           & 0.343          & 0.710          \\
PGD \cite{wang2022pgd}            & 1                  & 0.278           & 0.909           & 0.267           & 0.938           & 0.346           & 0.352          & 0.681          \\
PETR-R50 \cite{liu2022petr}       & 1                  & 0.225           & 0.859           & 0.314           & 0.862           & 0.271           & 0.376          & 0.605          \\
PETR-R101 \cite{liu2022petr}          & 1                  & 0.219           & 0.873           & 0.302           & 0.870           & 0.268           & 0.378          & 0.608          \\
PETR-Tiny \cite{liu2022petr}          & 1                  & 0.285           & 0.913           & 0.311           & 1.014           & 0.295           & 0.372          & 0.612          \\
BEVDet-Tiny \cite{bevdet2022}     & 1                  & 0.299           & 0.925           & 0.290           & 0.995           & 0.302           & 0.362          & 0.631          \\
BEVDet-Base \cite{bevdet2022}     & 1                  & 0.281           & 0.946           & 0.284           & 0.912           & 0.326           & 0.323          & 0.672          \\

\textbf{TopView (Proposed)}   & 1                  & 0.312           & 0.869           & 0.291          & 0.914           & 0.323           & 0.354         & 0.612          \\
\bottomrule
\end{tabular}
\end{table}

%%%%%%%%%%%%%%%%%%%%%%%%%%%%%%
%%%%%%%%%%%%%%%%%%%%%%%%%%%%%%%%

\subsection{Analysis, Discussion and Applications}

In this research, we introduce a new framework to transform CCTV video feeds into a live representation of objects and events in Google Maps that can be used for multipurpose urban analytics. We provide two major contributions: 1) We provide extensive analysis of London traffic at scale, detailing the contributions of each traffic mode to congestion during their various actions (i.e., standing or moving). 2) We provide a new approach for visualising CCTV data spatially and temporally. In doing so, we transform video data into a summary of lower-dimensional anonymized data that can be stored and retrieved with minimal memory and computational requirements. Google Maps' realistic approach may enable further spatial analysis using standard spatial methods directly from scaled maps.

\textbf{Generated trajectories and stream of paths:}

The introduced approach is not only useful for BEV map representation but also for representing high-dimensional streams of events of multifaced features as token objects with unique IDs over a given time interval (i.e., length of a given video file). Accordingly, this provides a summary of highly dimensional data such as video streams into ordered, easy to retrieve and anonymous data representation that is suitable for multi-purpose analysis. We showed multiple examples of how a video file can be transformed into a multi-dimensional vector representation, with road users represented as tokens in the video file's time interval, where each point in time of a given token carries information such as 3D bounding box, stationary status, class name, and so on.

\begin{figure}[h]
  \centering

   \includegraphics[width=1\linewidth]{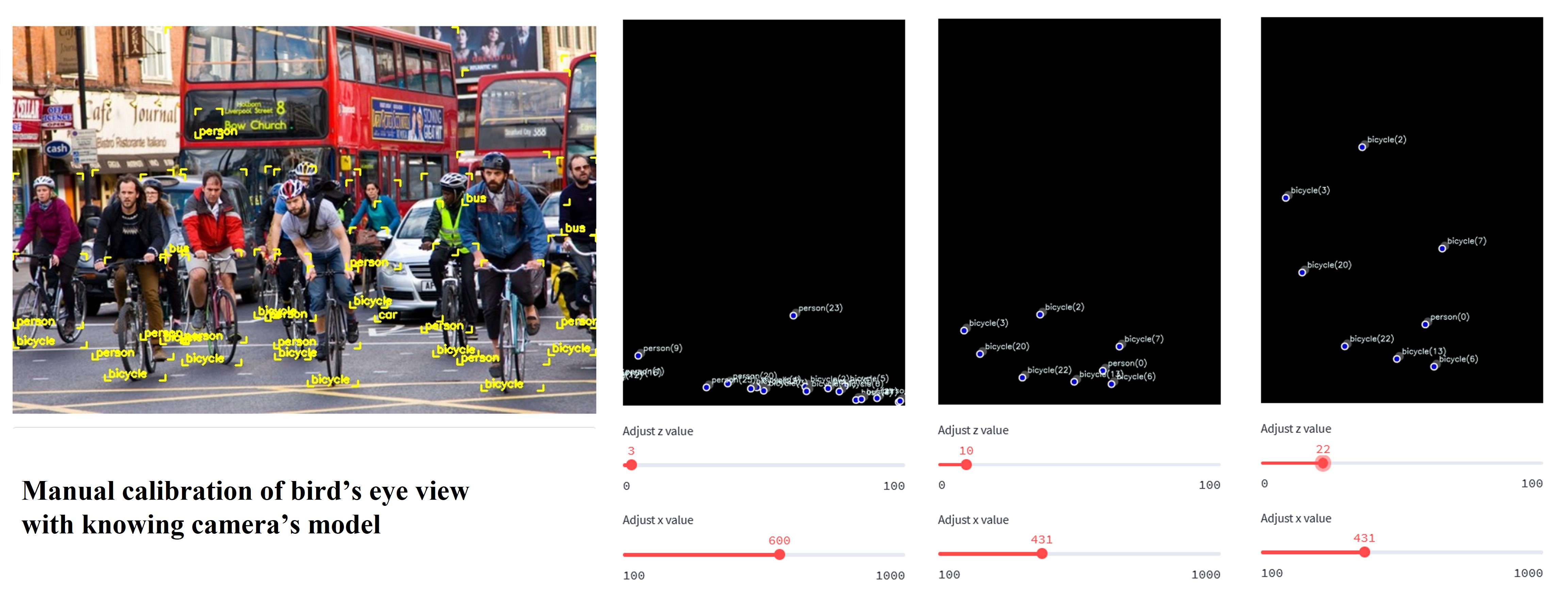}

   \caption{A manual calibration tool for adjusting the estimated bird’s eye view map from uncalibrated camera input such as internet images.}
   \label{fig:calibration}
\end{figure}

\textbf{ A simple manual calibration:}
Fig. \ref{fig:calibration} shows a simple tool for calibrating a bird’s eye map based on two values: 1) the z-value and 2) the x-value, without knowing the intrinsic and extrinsic parameters of a given camera. The z-value represents the spatial adjustment among the different objects in a given scene, and the x-value represents the shifting of road users in x-coordinates. While our framework generates an automatic bird's eye map, this tool provides additional control over the quality of the bird's eye map for manual calibration when necessary, particularly when linked to a Google Map.

\textbf{Application 1: High precision geo-localisation of road users and objects in a given scene:}
A direct application for the TopView framework is to localise and generate GPS trajectories of road users through a given camera’s location without knowing the GPS coordinates of individual road users. Here we show a few examples of localising road users through our framework in several CCTV cameras in London to show its versatility despite the complexity of road layouts. We showed a high precision in localising road users in the BEV map when compared to the road layouts of the camera feeds (See Fig. \ref{fig:fig1}). Despite the complexity of the street layout, we showed some examples of the estimated BEV map of road users at a top of a Google Map. It shows how accurately the model localises objects within a given street layout, paving the way for many applications that rely on data related to GPS trajectories or understanding the interactions among road users in a given scene.

\textbf{Application 2: Analysis of spatial occupancy of road and open spaces layouts:}
Understanding who uses which space of a given street, sidewalk or open space could be useful for several studies related to urban analytics and street design. Through the introduced framework, several studies can translate video streams of a given space to an occupancy map, highlighting the busiest spaces used by several road users, and spaces that are more likely to be deserted by road users. By doing so, current road layouts and open spaces can be evaluated and re-designed to meet the needs of their users based on their occupancy. In the future, pedestrian activities and actions can be analysed alongside their spatial occupancy to give empirical evidence of how spaces are used post-occupation.

\begin{figure}[H]
  \centering

   \includegraphics[width=1\linewidth]{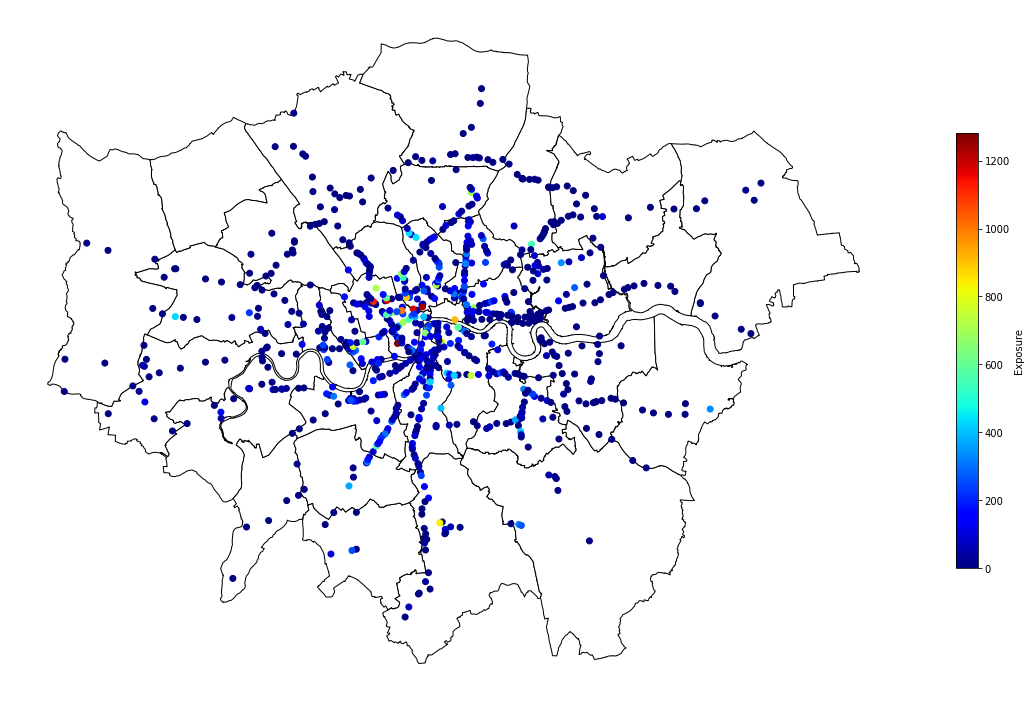}

   \caption{A map showing the level of exposure of pedestrians based on violating social distancing for a given hour $(20210813-0800_0900)$ as an application of the estimated BEV map. }
   \label{fig:london_map}
\end{figure}

\textbf{Application 3: Exposure based on violating social distancing at the city scale: } 
Another application of the introduced framework is to utilise the BEV map to analyse distances and safety-based thresholds among different road users to understand for instance, exposure, collisions, or even near misses. Here we utilised the BEV map to analyse distances between pedestrians with a two-metre threshold to show the level of exposure across London from 857 cameras at a given hour of a given day. Fig. \ref{fig:london_map} shows the count of human contacts that violate social distancing. Across all of London, the map shows that the majority of violations occurred in the city centre of London. This application shows the framework's versatility in shifting from observing several sites at a micro-level to a city scale.

\textbf{Limitations and future work: }
Several future investigations can be conducted to advance the introduced framework. First, the estimated 3D bounding boxes are based on moving objects, whereas it is limited when objects are stationary in a given scene. Further study can be done to rely on the relationship between the vanishing point of a given scene and the pose of a given object for stationary objects, instead of its trajectory line. Second, developing an automated method to transform scene components such as road layout \cite{source26, source27} and lane lines \cite{source52, source53} into a semantic vector representation would appear to be the next logical step in improving the introduced framework.

\section{Conclusion}
Scene awareness for a Multiview representation represents a crucial domain in vision and machine learning research. In this paper, we presented a hybrid method for estimating a vector representation of objects in a BEV without relying on the camera matrix information. This offers a similar approach to human navigation in spaces by understanding the relationship between objects rather than the exact depth of individual ones. Nevertheless, based on simple calibration, we also presented a geo-tagging of objects in a Google map with a very high spatial resolution which is useful for many applications related to urban analytics and autonomous navigation. Furthermore, this approach also provides a 3D bounding representation based solely on the Geometric transformation of the 2D bounding box and trajectory lines, in the case of sequential frames. This paper presents two opportunities for future research such as 1) 3D mesh representations of objects in complex scenes by learning from their multidimensional vector representation of point data, and 2) gamification of urban scene data and anonymising video stream data.

\section*{Conflict of Interest}
The author declares that there is no conflict of interest regarding the publication of this article.

\section*{Data Availability}
All raw data used in this study can be accessed online. The sources and methods for obtaining this data have been explained in the methodology section of this article.

\bibliography{sn-bibliography}% common bib file

\end{document}